\documentclass[letterpaper, 10 pt, conference]{ieeeconf}  %

\IEEEoverridecommandlockouts                              %

\usepackage{epsfig} %
\let\labelindent\relax
\usepackage{cite}
\usepackage{hyperref}
\hypersetup{
    colorlinks=true,
    linkcolor=black,
    urlcolor=blue,
    pdfpagemode=FullScreen,
    }

\usepackage{amsmath,amssymb,amsfonts}
\usepackage{bm}
\usepackage{algorithm}
\usepackage{algpseudocode}
\usepackage{mathtools}
\usepackage{graphicx}
\usepackage{textcomp}
\usepackage{xcolor}
\usepackage{booktabs} %
\usepackage{scalerel,stackengine}
\usepackage[per-mode=symbol]{siunitx}
\stackMath
\newcommand\reallywidehat[1]{%
\savestack{\tmpbox}{\stretchto{%
  \scaleto{%
    \scalerel*[\widthof{\ensuremath{#1}}]{\kern-.6pt\bigwedge\kern-.6pt}%
    {\rule[-\textheight/2]{1ex}{\textheight}}%
  }{\textheight}%
}{0.5ex}}%
\stackon[1pt]{#1}{\tmpbox}%
}

\usepackage{color}

\usepackage{changes}
\definecolor{maroon}{RGB}{128,0,128}
\definechangesauthor[name={M.~C.}, color={red}]{mc}
\definechangesauthor[name={S.~M.}, color={maroon}]{sm}

\title{\LARGE \bf Rapid and Reliable Quadruped Motion Planning \\ with Omnidirectional Jumping}

\usepackage{enumitem}

\newlist{SubItemList}{itemize}{1}
\setlist[SubItemList]{label={$-$}}

\let\OldItem\item
\newcommand{\SubItemStart}[1]{%
    \let\item\SubItemEnd
    \begin{SubItemList}[resume]%
        \OldItem #1%
}
\newcommand{\SubItemMiddle}[1]{%
    \OldItem #1%
}
\newcommand{\SubItemEnd}[1]{%
    \end{SubItemList}%
    \let\item\OldItem
    \item #1%
}
\newcommand*{\SubItem}[1]{%
    \let\SubItem\SubItemMiddle%
    \SubItemStart{#1}%
}%

\author{Matthew Chignoli\footnotemark$^{* 1}$, 
\thanks{$^{*}$Equal contribution.}
Savva Morozov\footnotemark$^{* 2}$, and Sangbae Kim\footnotemark$^{1}$%
\thanks{
$^{1}$Department of Mechanical Engineering and $^{2}$Department of Aeronautics and Astronautics, Massachusetts Institute of Technology, Cambridge, MA 02139, USA: {\tt\small \{chignoli,savva\}@mit.edu}}}%

\begin{document}

\maketitle
\thispagestyle{empty}
\pagestyle{empty}

\begin{abstract}
Dynamic jumping with legged robots poses a challenging problem in planning and control.
Formulating the jump optimization to allow fast online execution is difficult; efficiently using this capability to generate long-horizon motion plans further complicates the problem.
In this work, we present a hierarchical planning framework to address this problem.
We first formulate a real-time tractable trajectory optimization for performing omnidirectional jumping.
We then embed the results of this optimization into a low dimensional jump feasibility classifier.
This classifier is leveraged to produce geometric motion plans that select dynamically feasible jumps while mitigating the effects of the process noise.
We deploy our framework on the Mini Cheetah Vision quadruped, demonstrating the robot's ability to generate and execute reliable, goal-oriented plans that involve forward, lateral, and rotational jumps onto surfaces as tall as the robot's nominal hip height.
The ability to plan through omnidirectional jumping greatly expands the robot's mobility relative to planners that restrict jumping to the sagittal or frontal planes.

\end{abstract}

\newcommand{\Real}{\mathbb{R}}
\newcommand{\omegaBdHat}{\widehat{^B\bm{\omega}^d}}

\section{Introduction}

\begin{figure}
    \centering
    \includegraphics[width=\columnwidth]{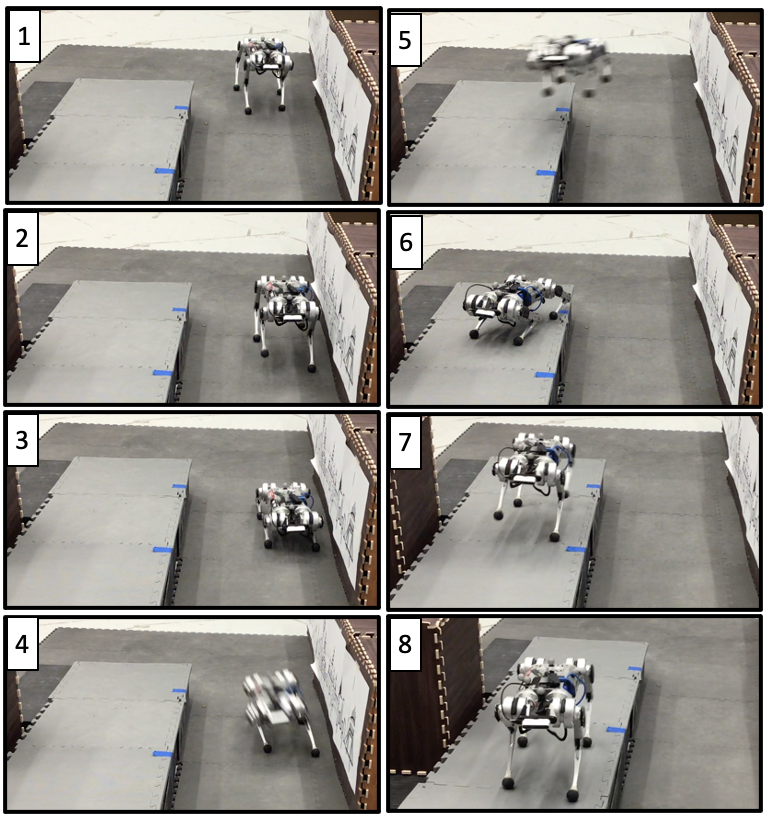}
    \caption{Mini Cheetah Vision executing a motion plan that involves a lateral jump onto a 20~\si{\centi\meter} surface.}
    \label{fig:mcv_jump_intro}
\end{figure}

Legged animals are capable of many agile movements beyond basic walking and running.
Jumping, especially omnidirectional jumping, allows robots to be deployed in environments with severely discontinuous elevation.
The ability to reliably execute these jumps, like the one shown in Fig.~\ref{fig:mcv_jump_intro}, is a challenge in its own right.
Planning when and how to jump adds another layer of complexity.
This planning should happen quickly---at a rate that matches the dynamism of the robot---and should be aware of the robot's physical capabilities to avoid planning infeasible jumps.
We formulate a trajectory optimization that efficiently generates omnidirectional jumps, and then incorporate the results of this optimization into a hierarchical navigation framework.
This allows us to rapidly plan motions through challenging environments while also mitigating the effects of process noise due to executing such motions on a physical robot.

\subsection{Related Work}

In model-based control frameworks, the complexity of the model typically scales with the difficulty of the task at hand.
In the case of omnidirectional jumping, especially jumps with large rotational components, the model needs to be suitably detailed to capture the complex dynamics. 
Trajectory optimization has proven to be a powerful tool for reasoning about such complicated models~\cite{posa2013direct, dai2014whole} and generating impressive aerial motions~\cite{zordan2014control, farshidian2017efficient, winkler2018gait}.
Yet performing these optimizations in real-time onboard a mobile robot is still challenging even for short-horizon motions~\cite{ponton2021efficient,ding2020kinodynamic}.

For tasks that require planning over longer time horizons, such as navigating through terrain with obstacles, complicated models are computationally intractable and at times unnecessary.
Consequently, hierarchical architectures are often implemented to decouple short and long-horizon planning tasks~\cite{li2020hybrid, wooden2010autonomous, brunner2013hierarchical,dinghybrid}.
Optimization-based methods, like those described above, tend to dominate lower hierarchy levels, while high-level planning can be accomplished using optimization-based~\cite{kwon2020fast} or sampling-based methods ~\cite{geisert2019contact,bouman2020autonomous, dang2020graph, fernbach2017kinodynamic} with lower-dimensional models. 

\begin{figure*}[t]
    \centering
    \includegraphics[width=500pt]{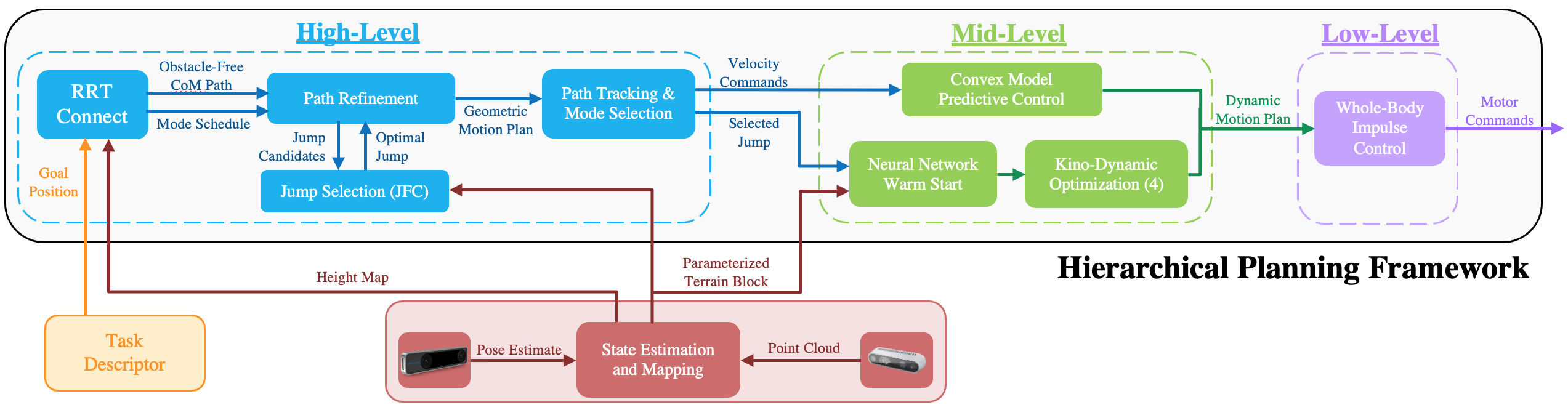}
    \caption{Overview of the proposed hierarchical navigation framework.}
    \label{fig:framework_bd}
\end{figure*}

Varying model complexity at different levels of the hierarchy creates a challenging trade-off.
If the reduced-order model is a conservative approximation of the robot's dynamics, a planner may fail to find a solution when one exists.
On the other hand, a lazy planner may produce a path that turns out to be infeasible when a more complete dynamics model is considered.
To address this trade-off, techniques for embedding whole-body dynamics into reduced-order models have been proposed, but are limited to simulation~\cite{norby2020fast,lin2019efficient} or to slow-moving, quasi-static motions~\cite{carpentier2018multicontact}.

Additionally, process noise due to external disturbances, imprecise path tracking, and other unmodeled dynamics is inevitable when deploying control frameworks on hardware.
Unless accounted for, this uncertainty may result in a failure mode, such as the robot colliding with obstacles.
A well-designed planning framework should then take into account the process noise that the robot will experience over the course of accomplishing its task.

\subsection{Contribution}

The contribution of this paper is a hierarchical planning and control framework that enables execution of motion plans that traverse complex multi-layered terrain.
To address the challenge posed by using models of varying complexity at different hierarchy levels, we learn a condensed representation of robot's jumping capabilities.
This classifier allows us to generate long-horizon, geometric motion plans that are dynamically feasible.
To account for the process noise in the hardware execution of these plans, we use a robustness metric to select jumps that minimize the risk of failure~\cite{heim2020learnable}.
We demonstrate the framework on the Mini Cheetah Vision~\cite{kim2020vision}. 
The robot successfully generates and follows goal-oriented motion plans that involve forward, lateral, and rotational jumps onto surfaces as high as the robot's nominal hip height.

The remainder of this paper adheres to the following structure.
Section II provides an overview of the framework.
Section III details the formulation of the kino-dynamic optimization that generates the jumping motions.
Section IV discusses the jump feasibility classifier used by the high-level planner and how variability in the jump parameters, driven by the process noise in hardware execution, informs jump selection.
Section V describes the planning algorithm used to generate reliable motion plans involving walking and jumping.
Section VI discusses the results of deploying this framework on the MIT Mini Cheetah Vision, and Section VII concludes the work and looks towards future research. 

\section{Framework}

In this section, we provide a brief overview of our hierarchical planning framework that enables navigation of challenging environments that require omnidirectional jumping.
First, the robot uses its perception stack (red block in Fig.~\ref{fig:framework_bd}) to estimate its pose and build a robot-centric elevation map of the environment.
The robot's operator then provides a desired position to the robot (orange block).
Taking into account the terrain elevation, the physical limits of the robot, and process noise of path tracking on hardware, the high-level sampling-based planner (blue block) generates a geometric motion plan.
This plan is comprised of a collision-free path together with a mode schedule of feasible actions that traverse this path.
A high-level tracking controller uses this data to issue the commands to the mid-level locomotion module (green block), which consists of walking and jumping controllers.
Walking is accomplished with a hybrid convex model predictive control/whole-body impulse control scheme, discussed in \cite{kim2019highly}.
The jumping controller, discussed in detail in Section III, is based on kino-dynamic trajectory optimization. 
The mid-level locomotion module determines the reaction forces and reference foot step locations, which are mapped to torque commands by the low-level whole body controller (purple block).
Our framework enables the robot to quickly plan and follow motion plans that cannot be generated under common simplifications like planar jumping or quasi-static walking.

\section{Kino-Dynamic Jump Optimization}

Online jumping optimizations are formulated as kino-dynamic motion planning problems. Kino-dynamic optimization simultaneously optimizes over the centroidal dynamics and the joint-level kinematics of the robot~\cite{dai2014whole}.
Such an optimization is well-suited for online 3D motions because the centroidal dynamics of the robot~\cite{orin2013centroidal} are both computationally simple to work with as well as expressive enough to produce diverse motions.
The kinematic aspect of the formulation is important because it ensures collision avoidance with the environment.

The centroidal dynamics of the robot can be described by state and control trajectories,
\begin{equation}
    \mathbf{X}^\text{dyn}(t) = \begin{bmatrix} \mathbf{r}(t) \\ \mathbf{\dot{r}}(t) \\ \mathbf{h}(t) \\ \mathbf{c}_1(t) \\ \vdots \\ \mathbf{c}_{n_c}(t) \end{bmatrix}, \quad \mathbf{U}^\text{dyn}(t) = \begin{bmatrix} \mathbf{f}_1(t) \\ \vdots \\ \mathbf{f}_{n_c}(t) \end{bmatrix} \label{eq:Xd}
\end{equation}
where $\mathbf{r}\in\Real^3$ is the center of mass trajectory, $\mathbf{\dot{r}}\in\Real^3$ is the center of mass velocity trajectory, $\mathbf{h}\in\Real^3$ is the centroid angular momentum, and $\mathbf{c}\in\Real^{3n_{c}}$ and $\mathbf{f}\in\Real^{3n_{c}}$ are the contact locations and ground reaction forces at the $n_c$ contact points, respectively. 
The kinematic trajectory of the robot is given by
\begin{equation}
    \mathbf{X}^\text{kin} = \begin{bmatrix} \mathbf{q}(t) \\ \mathbf{\dot{q}}(t) \end{bmatrix} \label{eq:Xk}
\end{equation}
where, for a robot with $n$ degrees of freedom (DoF), including a six DoF floating base, $\mathbf{q}\in\Real^n$ are the joint positions and $\mathbf{\dot{q}}\in\Real^n$ are the joint velocities. 

To formulate the problem as a trajectory optimization, the state and control trajectories are discretized into a set of $N_{TO}$ timesteps corresponding to the takeoff period (all legs in contact) and a set of $N_{FL}$
timesteps corresponding to the flight period (all legs out of contact).
The total number of timesteps for the optimization is, therefore, $N=N_{TO}+N_{FL}$.
The duration of the takeoff period, $T_{TO}$, is chosen beforehand, while the duration of the flight phase, $T_{FL}$, is left as an optimization variable.

After discretization, the optimization variables \eqref{eq:Xd} and \eqref{eq:Xk} become
\begin{equation}
    \begin{split}
        \mathbf{X}^\text{dyn} &= \begin{bmatrix} \mathbf{X}^\text{dyn}_1, \ldots, \mathbf{X}^\text{dyn}_{N} \end{bmatrix}, \\
        \mathbf{U}^\text{dyn} &= \begin{bmatrix} \mathbf{U}^\text{dyn}_1, \ldots, \mathbf{U}^\text{dyn}_{N_{TO}-1} \end{bmatrix}, \\
        \mathbf{X}^\text{kin} &= \begin{bmatrix} \mathbf{X}^\text{kin}_1, \ldots, \mathbf{X}^\text{kin}_{N} \end{bmatrix}.
    \end{split}
\end{equation}

The goal of the trajectory optimization proposed in this work is to find a control policy $\mathbf{U}^\text{dyn}$ that results in the robot safely jumping onto an elevated surface.
We formulate this optimization according to
\begin{subequations}
\begin{align}
    \min_{\mathbf{X}^\text{dyn},\mathbf{U}^\text{dyn},\mathbf{X}^\text{kin}}& \big\vert\big\vert \mathbf{\tilde{X}}^\text{dyn} \big\vert\big\vert^2_{Q_d} + \big\vert\big\vert \mathbf{\tilde{X}}^\text{kin} \big\vert\big\vert^2_{Q_k} + \big\vert\big\vert \mathbf{U}^\text{dyn} \big\vert\big\vert^2_{Q_u} \label{eqn:cost} \\
    \text{s.t.            } & \mathbf{X}^\text{dyn}_{i+1} = \mathbf{X}^\text{dyn}_{i} + f_\text{dyn}(\mathbf{X}^\text{dyn}_i,\mathbf{U}^\text{dyn}_i)\Delta t_i \label{eqn:dyn_cnstr} \\
    & \mathbf{q}_{i+1} = \mathbf{q}_{i} + \dot{\mathbf{q}}_i\Delta t_i \label{eqn:kin_cnstr} \\
    & \mathbf{X}^\text{dyn}_i = \text{kinematics}(\mathbf{X}^\text{kin}_i) \label{eqn:agree_cnstr} \\
    & \mathbf{U}^\text{dyn}_i \in \mathcal{U}, \quad \mathbf{c}_i \in \mathcal{C}_i \label{eqn:contact_cnstr} \\
    & \mathbf{X}^\text{kin}_1 = \mathbf{X}^\text{kin}(0), \quad \mathbf{X}^\text{dyn}_N \in \mathcal{X}_\text{land} \label{eqn:bound_cond}
\end{align}
\end{subequations}
The cost function~\eqref{eqn:cost} penalizes deviations from heuristically derived reference trajectories $\mathbf{X}^\text{dyn}_\text{ref}$ and $\mathbf{X}^\text{kin}_\text{ref}$, where
\begin{equation}
    \mathbf{\tilde{X}}^\text{dyn} = \mathbf{X}^\text{dyn} - \mathbf{X}^\text{dyn}_\text{ref}, \quad \mathbf{\tilde{X}}^\text{kin} = \mathbf{X}^\text{kin} - \mathbf{X}^\text{kin}_\text{ref},
\end{equation}
and $Q_d$, $Q_k$, and $Q_u$ are the weighting matrices. The heuristic references are ballistic trajectories for the CoM and linear interpolations between the initial and final orientation of the robot.

The centroidal dynamics of the system are encoded in~\eqref{eqn:dyn_cnstr} via
\begin{equation}
    \begin{bmatrix} \ddot{\mathbf{r}} \\ \dot{\mathbf{h}} \end{bmatrix} = \begin{bmatrix} \frac{1}{m}\sum_{k=1}^{n_c}\mathbf{f}_{k,i} \\ \sum_{k=1}^{n_c}(\mathbf{c}_{k,i}-\mathbf{r}_i)\times\mathbf{f}_{k,i} \end{bmatrix},
\end{equation}
where $m$ is the mass of the robot and $\mathbf{f}_{k,i}\in\Real^3$ is the reaction force at the $k$th contact point at the $i$th timestep.
Note that for~\eqref{eqn:dyn_cnstr} and~\eqref{eqn:kin_cnstr}, the timestep $\Delta t_i$ is different for the takeoff and flight phases,
\begin{equation}
    \Delta t_i =
\begin{cases}
    \frac{T_{TO}}{N_{TO}-1},& i<N_{TO}\\
    \frac{T_{FL}}{N_{FL}-1}, & N_{TO}\leq i< N,
\end{cases}
\end{equation}
since $T_{TO}$ is fixed while $T_{FL}$ is an optimization variable.
This added nonlinearity to~\eqref{eqn:dyn_cnstr} and~\eqref{eqn:kin_cnstr} is necessary because we cannot know a priori how long the flight should last.
Underestimation will not give the robot enough time to reach the required height, while overestimation will cause the robot to expend unnecessary energy or, worse yet, call for an overly aggressive jump that exceeds the robot's torque limits.

Agreement between the centroidal dynamic and the kinematic states of the robot is enforced via~\eqref{eqn:agree_cnstr}.
This constraint ensures that at each time step, applying forward kinematics to $\mathbf{X}^\text{kin}_i$ will yield a CoM and contact point locations that match $\mathbf{X}^\text{dyn}_i$.
Furthermore, it enforces that the total angular momentum of the individual links of the robot is equal to the centroidal angular momentum~\cite{orin2013centroidal}.

Contact constraints are enforced via~\eqref{eqn:contact_cnstr}.
The set of feasible ground reaction forces $\mathcal{U}$ is given by a friction pyramid inscribed in the cone
\begin{equation}
    \bigg\{(f_x,f_y,f_z)\in\Real^3 \bigg\vert \sqrt{f_{x}^2+f_{y}^2} \le \mu f_{z}\bigg\},
\end{equation}
where $\mu$ is the coefficient of friction with the ground.
The set of allowable footstep locations $\mathcal{C}$ depends on the phase.
In takeoff, i.e., for $i\le N_{TO}$, $\mathcal{C}_i$ enforces that the feet remain at their original position.
In flight, $\mathcal{C}_i$ is the set of all footstep positions above the terrain
\begin{equation}
    \mathcal{C}_i = \bigg\{(c_x,c_y,c_z)\in\Real^3 \bigg\vert c_z > z_g(c_x,c_y)\bigg\}, \quad N_{TO}\le i < N,
\end{equation}
where $z_g$ is a continuous function that maps the $x-y$ position of the foot to the height of the terrain at that point.
In our case, we approximate discrete changes in elevation using the sigmoid function
\begin{equation}
    z_g(x,y) = \frac{a}{1+e^{-\gamma(x-b-dy)}}
\end{equation}
where $a$, $b$, and $d$ are the parameters fit from the height map data, and $\gamma$ is a manually selected parameter that dictates the steepness of the sigmoid.

Lastly, boundary conditions for the state of the robot are given by~\eqref{eqn:bound_cond}: $\mathbf{X}^\text{kin}(0)$ is the initial state and $\mathcal{X}_\text{land}$ is the set of allowable landing states on the elevated surface.
The values associated with $\mathbf{X}^\text{kin}(0)$ come from the robot's state estimator, while the values associated with $\mathcal{X}_\text{land}$ come from the high-level path planner.

\section[Feasibility Classifier: Selecting Reliable Jumps]{Feasibility Classifier: \\Selecting Reliable Jumps}

For the high-level planner to generate long-horizon motion plans through walking and jumping, we need a condensed representation of the robot's jumping capabilities.
We first define a low-dimensional parameterization of the kino-dynamic jumping optimization (Section~\ref{ssec:jump_parameterization}).
We then train a classifier, hereafter referred to as the ``jump feasibility classifier" (JFC), that takes in the five-dimensional parameterization of an intended jump and outputs a binary classification of the jump as feasible or infeasible.
This offers a computationally efficient way to select viable jumps, ensuring that feasible high-level motion plans can be generated quickly.
We describe the JFC training procedure in Section~\ref{ssec:jfc_train}.
Finally, Section~\ref{ssec:max_robust} explains how the JFC also offers a way to select jumps that are robust to process noise.

\subsection{Low-dimensional Jump Parameterization} \label{ssec:jump_parameterization} 
\begin{figure}
    \centering
    \includegraphics[width=\columnwidth]{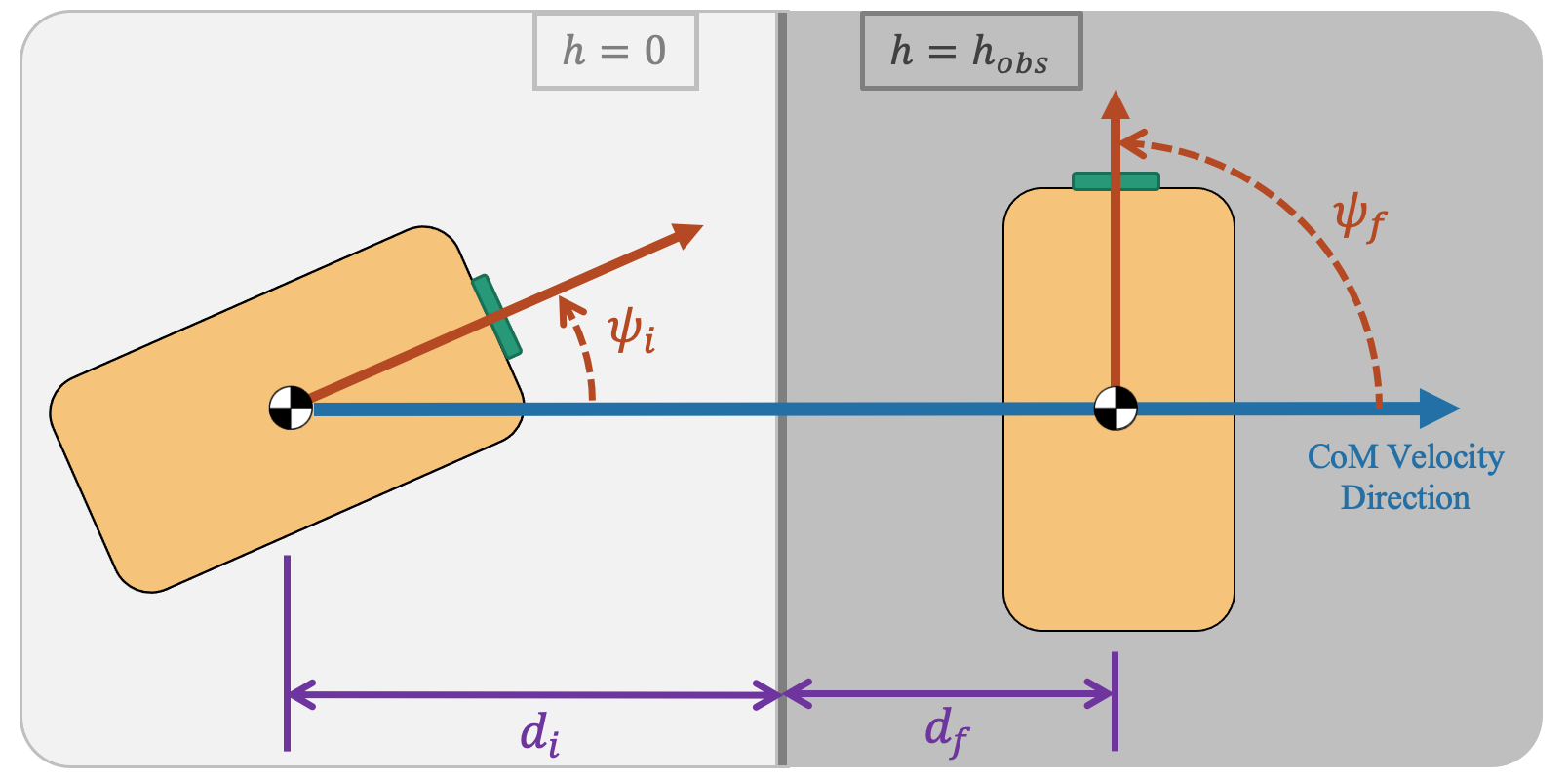}
    \caption{Overhead view of the robot before (left) and after (right) it jumps onto an elevated surface, along with the five-dimensional space of parameters that define the jump---$h_{obs},d_i,\psi_i$, $d_f$, $\psi_f$.}
    \label{fig:jump_param}
\end{figure}

For tractability reasons, we make the following simplifications regarding the jumps performed by the robot: takeoff and landing terrain must be flat, the jump surface edges are straight lines (i.e., vertical walls), and the robot's CoM velocity is perpendicular to the surface it is jumping onto.
These assumptions allow us to describe any jump by a point in a 5D parameter space, illustrated in Fig~\ref{fig:jump_param}.
Despite the low dimensionality of the parameter space, it still captures a wide range of potential jumping motions.
The parameters are as follows: $h_{obs}$ is the height difference between the surfaces, $d_i$ is the initial distance from the CoM to the surface edge, $\psi_i$ is the robot's initial yaw relative to the edge, $d_f$ is the final distance of the CoM past the surface edge, and $\psi_f$ is the robot's final yaw.

\subsection{Jump Feasibility Classifier Training} \label{ssec:jfc_train}

We train the JFC from a set of 12,000 offline jump simulations uniformly sampled from the 5D space of jumping parameters.
For each sample, we first carry out the corresponding jump optimization as in (4).
Using this optimal trajectory, the robot attempts to perform the jumping motion in simulation.
We classify successful jumps as ones where the robot ends with its CoM at the desired height and with an upright posture (i.e., the robot has not fallen over). 
Any other outcome is classified as a failure.
We use the collected data, which maps jump parameters to a binary outcome, to train a support vector machine (SVM). 
This SVM is a compact way to encode the robot's observed jumping capabilities into a classifier that can be evaluated efficiently onboard the robot as it refines high-level paths.

\subsection{Identifying Maximally Robust Jumps} \label{ssec:max_robust}

Given a continuous 5D space of jump parameters $\mathcal{J}$, we aim to find the jump that is maximally robust to the hardware process noise.
To pick such jump, we first constrain $\mathcal{J}$ to only include the parameters classified as feasible by the JFC. 
This produces the set of \textit{physically} feasible jump parameters $\mathcal{J}_{\text{P}}$, i.e., jumps that the robot is capable of accomplishing.
Next, we impose the terrain collision avoidance constraints, which produces the \textit{terrain}-constrained set of parameters $\mathcal{J}_{\text{T}}$.
The intersection of $\mathcal{J}_{\text{P}}$ and $\mathcal{J}_{\text{T}}$ is the \textit{feasible action} set $\mathcal{J}_{\text{FA}}$: any point in $\mathcal{J}_{\text{FA}}$ parameterizes a jump that is both physically feasible and collision-free with respect to the terrain.
Finally, we select the point that is furthest from the $\mathcal{J}_{\text{FA}}$ set boundary, as this results in the largest safety margin.
We find this point using a pole of inaccessibility (PIA) search algorithm similar to~\cite{garcia2007poles}.

\section{High-Level Planning}

The goal of the high-level planner is to generate geometric motion plans that are feasible and collision-free.
The high-level planner operates in a cascaded manner: we first use a sampling-based planner to find a coarse collision-free path to the goal (Section~\ref{ssec:sampling-coarse-plan}), and then refine this path to ensure dynamic feasibility of the jumps within it (Section~\ref{ssec:motion-plan-refinement}).

\subsection{Onboard Perception and Mapping}
\label{ssec:perception-and-mapping}
Onboard perception, mapping, and state estimation are handled via a suite of Intel RealSense cameras.
Terrain data is obtained as a point cloud from the D435 camera and is inserted into a height map~\cite{fankhauser2016universal}.
This robot-centric height map is propagated using a 200Hz pose estimate produced by the T265 camera.
The terrain is subsequently filtered and segmented into flat surfaces~\cite{ester1996density}.
In this work, we manually map the terrain before each test by pitching and yawing the robot; perception-aware planning, while preferred, is outside the scope of this paper.

\begin{figure*}[t!]
    \centering
    \includegraphics[width=450pt]{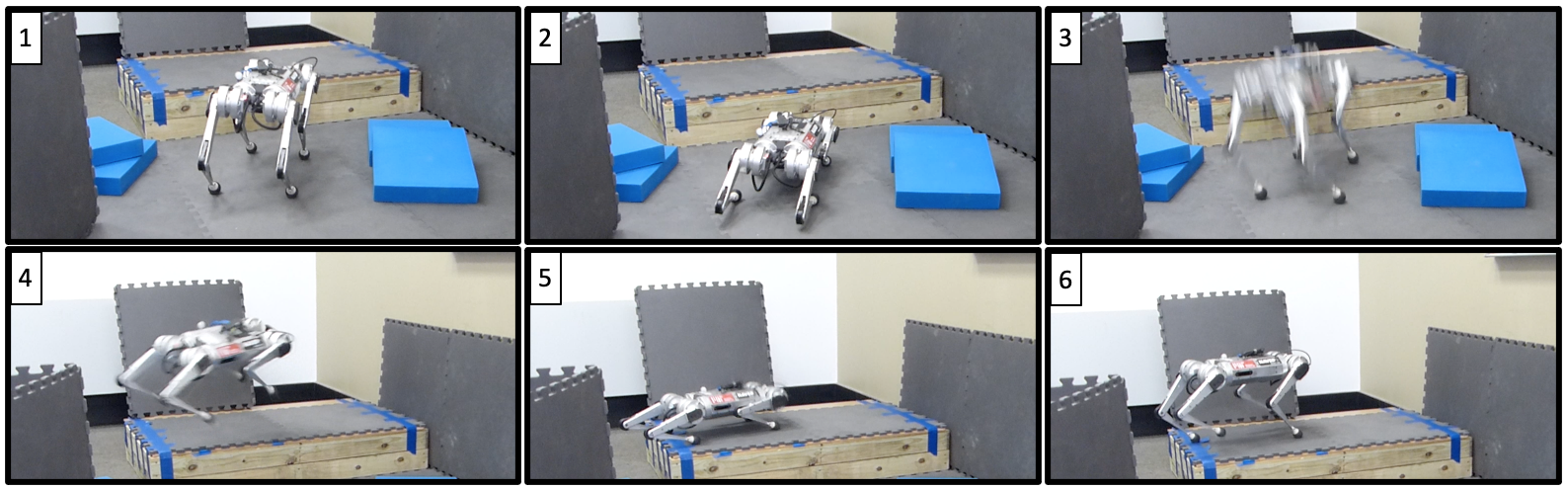}
    \caption{Mini Cheetah Vision executing a rotational jump onto 22~\si{\centi\meter} jump surface.}
    \label{fig:rotational}
\end{figure*}

\subsection{Sampling-Based Coarse Path Planning}
\label{ssec:sampling-coarse-plan}
Given the terrain, the current robot's pose, and the desired location of the robot, we use RRT-connect~\cite{kuffner2000rrt} to find a collision-free CoM path to the goal.
This coarse path is geometric: it does not consider the robot's dynamics and uses relaxed traversability heuristics to quickly obtain a rough motion plan estimate.
Two heuristics are used: path segments that belong to a single flat surface are considered walkable; the segments between different flat surfaces are accepted as jumps if the surface height difference is within a 25cm threshold and if both takeoff and landing locations are collision-free.

\subsection{Motion Plan Refinement and Tracking} %
\label{ssec:motion-plan-refinement}
In this stage, we refine the original, coarse geometric path into a reliable and dynamically feasible motion plan.
Segments of the coarse path that correspond to walking are modified by locally moving the path vertices away from the obstacles using gradient descent.
The high level tracking controller, similar to~\cite{wooden2010autonomous}, is used to track the walking portions of the motion plan, while also attempting to keep the robot away from the obstacles.
We presume that the tracking and locomotion controllers can produce a collision-free realization of the walking segments, but this is not guaranteed.
The effect of this assumption is mitigated by the fact that the degrees of freedom of the robot's CoM are independently controllable in the transverse plane.

The coarse path does not specify a particular jump for the robot to execute---rather, it constrains the space of feasible jump parameters.
Specifically, the terrain around the proposed jump location constrains the initial $d_i$ and $\psi_i$ and the terminal $d_t$ and $\psi_t$ to ensure that the takeoff and landing are collision-free, and also constrains the obstacle height $h_{obs}$ to a particular value.
The most robust jump is then selected according to the process described in Section~\ref{ssec:max_robust}.
If this refinement of the coarse path into a dynamically feasible motion plan is impossible (i.e., the set $\mathcal{J}_{\text{FA}}$ is empty), then a new coarse path is generated, and the process is repeated.

\section{Results}

In this section, we present the hardware implementation of our hierarchical planning framework.
We first provide solve-time statistics for the underlying planners to demonstrate the framework's capacity to run online on a robot.
We then discuss our ability to perform omnidirectional jumps and how this expands the robot's capabilities.
Lastly, we visualize the process of selecting maximally robust jump parameters and further motivate this design choice. 
A full demonstration of our proposed framework is included in the \href{https://youtu.be/gGVENdcfetQ}{attached video}.

\subsection{Hardware Setup}
The entire framework runs in real-time onboard a robot with only modest computational capabilities.
We deploy our framework on a robot equipped with two onboard computers: an NVIDIA Jetson TX2 and an Intel UP Board.
The more powerful TX2 is used for the more computationally intensive processes (perception, mapping, high-level planning, and nonlinear trajectory optimization). 
The UP Board handles simple, lower-level tasks such as locomotion control and communication with the robot's motor controllers.

\subsection{Fast Online Planning}

A major feature of the proposed framework is its fast onboard computation time.
To reduce the solve time of our trajectory optimization, we use the technique of warm starting our optimization using a neural network trained from offline data~\cite{melon2020reliable,surovik2020learning}.
Jump optimizations~(4) were solved on the TX2 using a commercial nonlinear solver KNITRO~\cite{byrd2006k}. 
The average solve time was $1.91$~\si{\second} (standard deviation $0.43$~\si{\second}).
While this solve rate is too slow to be used for receding-horizon control, it is faster compared to other schemes that target this versatile behavior of omnidirectional jumps.
These optimizations are typically so computationally expensive that they must be either computed offline~\cite{dai2014whole} or on computers too bulky to be placed onboard the robot~\cite{ponton2021efficient}.

After 50 experiments over various terrain configurations, we observe that our high-level motion planner, which typically generates paths between 300-500~\si{\cm} long, takes on average $0.55$~\si{\second} to find a plan (standard deviation $0.32$~\si{\second}).
Since the robot is moving at approximately $0.25$~\si{\meter\per\second} for the duration of the path, the planner can, on average, generate motion plans at a rate $30$ times faster than it takes to execute them.

\subsection{Planning with Omnidirectional Jumping}

The ability to perform omnidirectional jumping greatly expands the set of terrains the robot can traverse compared to a system that makes conventional assumptions such as planar jumping. 
Consider the lateral jumping motion animated in Fig.~\ref{fig:mcv_jump_intro}.
If the robot tried to turn and face the jump surface head-on---a commonly used heuristic---it would collide with either the jump surface on the left or the wall on the right.

\begin{figure}
    \centering
    \includegraphics[width=185pt]{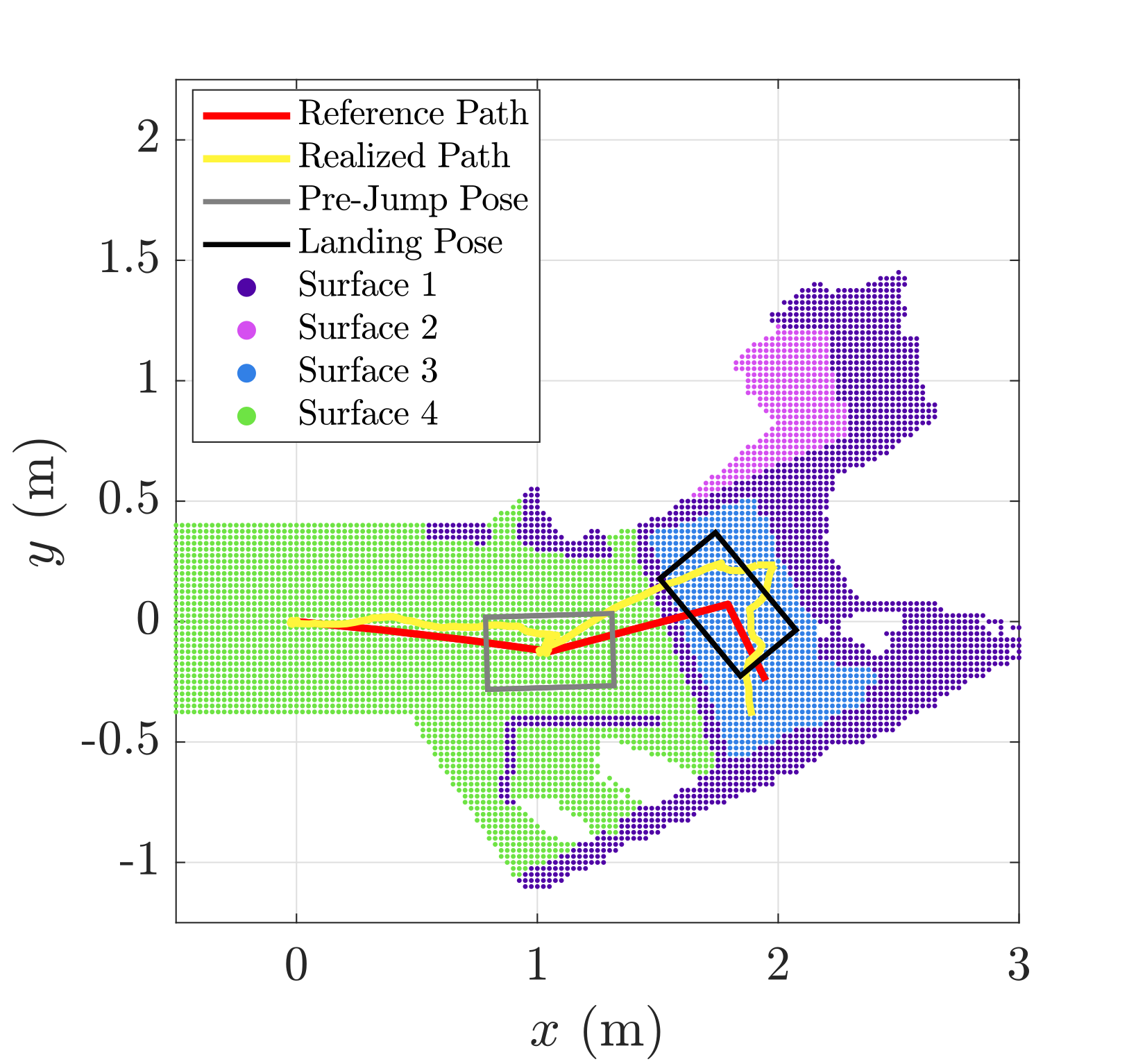}
    \caption{Planned and realized paths of the robot as it traverses terrain that requires a rotational jump onto a surface.}
    \label{fig:heightmap}
\end{figure}

This point is further emphasized by the rotational jump animated in Fig.~\ref{fig:rotational}. 
The CoM trajectory and the terrain height map are illustrated in Fig.~\ref{fig:heightmap}. 
The constraints of this setting are such that the robot must rotate anywhere from 60$^\circ$ to 100$^\circ$ while in the air to produce a safe, collision-free landing.

The terrain in Fig.~\ref{fig:heightmap} offers difficult constraints on the aerial rotation during the jump: to avoid collision, the takeoff yaw must be between -30$^\circ$ and 20$^\circ$, and the landing yaw between -100$^\circ$ and -70$^\circ$.
Common heuristics are insufficient to traverse this terrain. 
Indeed, purely sagittal or frontal plane jumps would fail to find a solution or produce a terrain collision. 
It is also incorrect to assume that all collision-free jumps are executable, since many of them are not dynamically feasible.
JFC allows us to efficiently select feasible jumps as a function of the terrain conditions and robot's capabilities.

\subsection{Robust Jumping: Planning Under Uncertainty}

\begin{figure}
    \centering
    \includegraphics[width=\columnwidth]{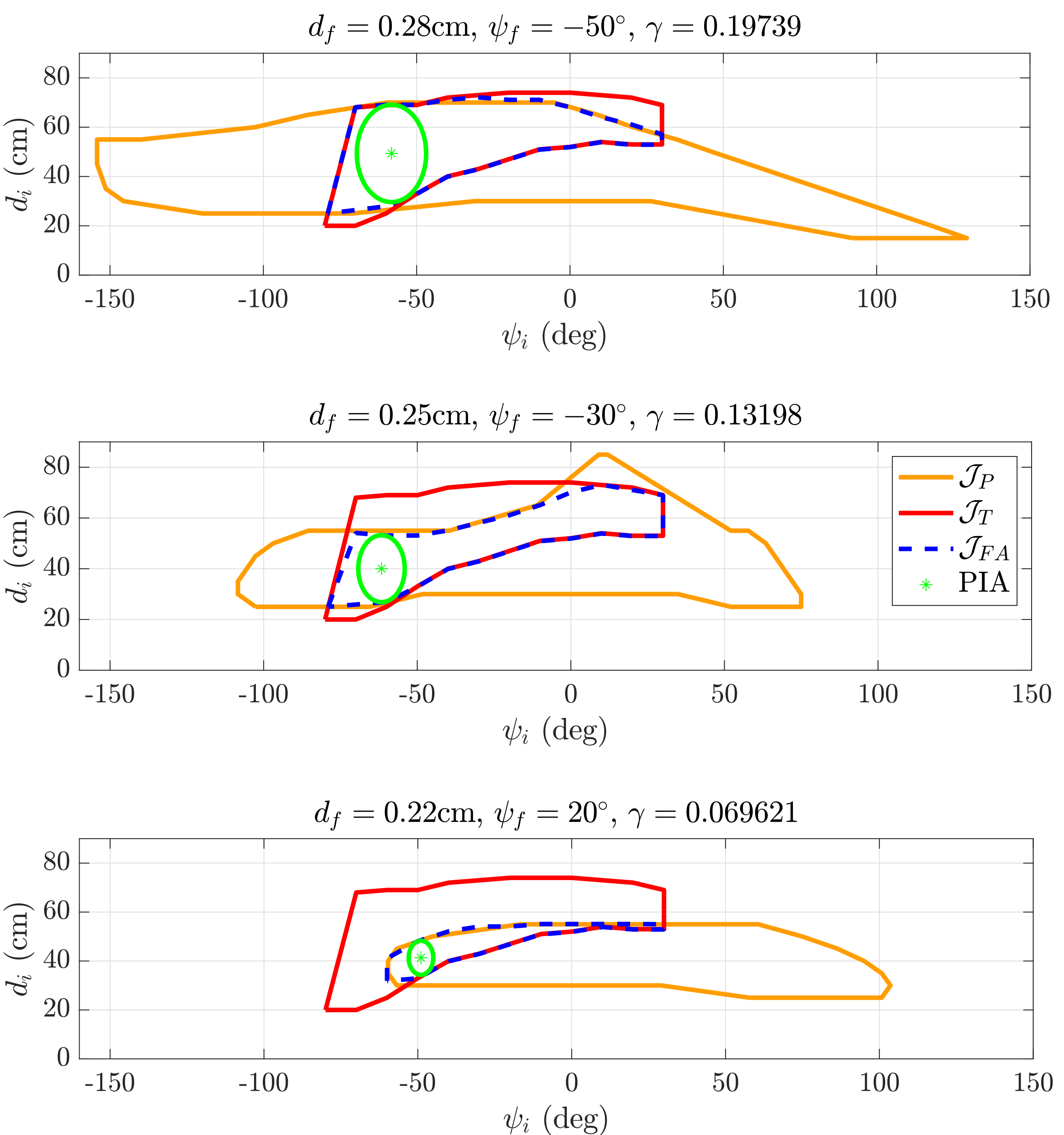}
    \caption{Two-dimensional slices of the intersection of the terrain-constrained set $\mathcal{J}_{\text{T}}$ and the physically feasible parameter set $\mathcal{J_{\text{P}} }$. 
    For each slice, the PIA of the intersection $\mathcal{J}_{\text{FA}}$ is highlighted. 
    The radius of the largest inscribed circle centered at the PIA is given by $\gamma$.}
    \label{fig:feas_polys}
\end{figure}

The process of finding maximally reliable jumps is examined in Fig.~\ref{fig:feas_polys} (for plotting simplicity, we consider fixed height and terminal states here).
The orange polygon, obtained from the offline-generated JFC, captures the range of initial jump conditions that the robot is physically capable of performing, $\mathcal{J}_{\text{P}}$.
The red polygon, obtained online in one of our hardware experiments, represents the terrain-constrained set of jump parameters, $\mathcal{J}_{\text{T}}$.
The intersection of $\mathcal{J}_{\text{P}}$ and $\mathcal{J}_{\text{T}}$ is depicted in blue.
This intersection is the $\mathcal{J}_{\text{FA}}$, and its PIA, shown in green, represents the point that is maximally robust to the variability in the initial conditions of the jump.
By performing this process in the 5D space of jump parameters, we are able to obtain a jump that can handle the largest deviation between the desired trajectory and its realization.

This example illustrates the advantage of our framework.
Simply picking any collision-free jump from $\mathcal{J}_{\text{T}}$ (red) is insufficient, since for some terminal conditions $(d_f,\psi_f)$, most of the jumps in $\mathcal{J}_{\text{T}}$ are not dynamically feasible---see the bottom plot of Fig~\ref{fig:feas_polys}.
Furthermore, picking an arbitrary point from $\mathcal{J}_{\text{FA}}$ may also be insufficient, as perturbations in the jumping parameters may result in a collision or render the jump optimization infeasible.
By finding the PIA of the entire 5-dimensional set $\mathcal{J}_{\text{FA}}$, we ensure that the selected jump is maximally robust to the parameter variability caused by the process noise during execution.

\section{Conclusions \& Future Work}

This work first presented a trajectory optimization formulation for generating online, omnidirectional jumps. 
The jumping capabilities endowed by this trajectory optimization are then leveraged by a high-level geometric motion planner.
This planner considers the physical limits of the robot as well as external constraints of the terrain to select maximally reliable jumps that navigate the robot to a goal position.
We synthesize the two planners into a hierarchical framework that allows the Mini Cheetah Vision to rapidly generate and complete reliable plans through challenging terrain.

In the future, we aim to improve on our work in a number of ways.
First, perception and mapping were performed manually in this work, and we are looking into observability-based planning to resolve this.
Second, measurement uncertainty was not accounted for in the JFC and is something we want to consider in future formulations.
Finally, we will focus on leveraging the JFC to produce provable robustness guarantees.
Currently, we select a jump that is maximally robust to parameter variability given a candidate motion plan.
This practical approach ensures that if a set of safe jumps does exist, our algorithm will have found one such jump. 
However, we do not make any claims about the existence of such a safe set---providing certificates of robust plan execution is beyond the scope of this paper but a goal for future work.

\section*{Acknowledgments}
This work was supported by NAVER Labs, Toyota Research Institute, the Centers for ME Research and Education at MIT and SUSTech, Boeing as part of the MIT SuperUROP program, and the National Science Foundation Graduate Research Fellowship Program under Grant No. 4000092301.

\bibliographystyle{IEEEtran}
\bibliography{icra2022_bib}

\end{document}